\documentclass[sn-mathphys, iicol]{sn-jnl}% Math and Physical Sciences Reference Style

\usepackage[utf8]{inputenc}
\usepackage{xcolor}
\usepackage{amssymb}
\usepackage{float}
\usepackage{amsmath}
\usepackage{hyperref}
\usepackage{graphicx}
\usepackage{caption}
\usepackage{subcaption}
\usepackage{booktabs} 
\usepackage{algorithm,algpseudocode}
\usepackage{array}

\begin{document}

\title[Precipitation nowcasting]{Precipitation nowcasting with generative diffusion models}

\author*[1]{\fnm{Andrea} \sur{Asperti}}\email{andrea.asperti@unibo.it}
\equalcont{These authors contributed equally to this work.}

\author*[1]{\fnm{Fabio} \sur{Merizzi}}\email{fabio.merizzi@unibo.it}
\equalcont{These authors contributed equally to this work.}

\author[1]{\fnm{Alberto} \sur{Paparella}}\email{alberto.paparella2@studio.unibo.it}
\equalcont{These authors contributed equally to this work.}

\author[2]{\fnm{Giorgio} \sur{Pedrazzi}}\email{g.pedrazzi@cineca.it}

\author[2]{\fnm{Matteo} \sur{Angelinelli}}\email{m.angelinelli@cineca.it}

\author[1]{\fnm{Stefano} \sur{Colamonaco}}\email{stefano.colamonaco@studio.unibo.it}


\affil*[1]{\orgdiv{Department of Informatics: Science and Engineering (DISI)}, \orgname{University of Bologna}, \orgaddress{\street{Mura Anteo Zamboni 7}, \city{Bologna}, \postcode{40126},
\country{Italy}
}}

\affil[2]{\orgdiv{HPC Department} 
\orgname{Cineca}, \orgaddress{\street{Magnanelli 6/3}, \city{Casalecchio di Reno (BO)}, \postcode{40033},
\country{Italy}
}}

%\affil[2]{\orgdiv{Department}, \orgname{Organization}, \orgaddress{\street{Street}, \city{City}, \postcode{10587}, \state{State}, \country{Country}}}

%\affil[3]{\orgdiv{Department}, \orgname{Organization}, \orgaddress{\street{Street}, \city{City}, \postcode{610101}, \state{State}, \country{Country}}}

\abstract{
In recent years traditional numerical methods for accurate weather prediction have been increasingly challenged by deep learning methods. Numerous historical datasets used for short and medium-range weather forecasts are typically organized into a regular spatial grid structure. This arrangement closely resembles images: each weather variable can be visualized as a map or, when considering the temporal axis, as a video. Several classes of generative models, comprising Generative Adversarial Networks, Variational Autoencoders, or the recent Denoising Diffusion Models have largely proved their applicability to the next-frame prediction problem, and is thus natural to test their performance on the weather prediction benchmarks. Diffusion models are particularly appealing in this context, due to the intrinsically probabilistic nature of weather forecasting: what we are really interested to model is the {\em probability distribution} of weather indicators, whose expected value is the most likely prediction.

In our study, we focus on a specific subset of the ERA-5 dataset, which includes hourly data pertaining to Central Europe from the years 2016 to 2021. Within this context, we examine the efficacy of diffusion models in handling the task of precipitation nowcasting. Our work is conducted in comparison to the performance of well-established U-Net models, as documented in the existing literature.
Our proposed approach of Generative Ensemble Diffusion (GED) utilizes a diffusion model to generate a set of possible weather scenarios which are then amalgamated into a probable prediction via the use of a post-processing network. 
This approach, in comparison to recent deep learning models, substantially outperformed them in terms of overall performance.
}

\maketitle
\noindent \textbf{keywords:} diffusion models, precipitation nowcasting, ensemble diffusion, weather forecasting, post-process

\section{Introduction}
The term {\em nowcasting} refers to the forecasting of weather indicators on a short-term meteo-scale period, typically between 2 and 6 hours. The forecast is an extrapolation in time of known weather parameters, comprising information obtained by means of remote sensing, radar echoes, and satellite data.

While numerical weather prediction models may accurately forecast the likelihood and overall intensity of precipitation across large geographical areas and medium-term temporal intervals, the situation is more challenging when it comes to short spatial and temporal scales \cite{AStudyontheScale}: on less than 4 hours predictions, they frequently perform worse than persistence-based forecasts \cite{ashok2022systematic}. This is mostly due to the high stochastic nature of the phenomenon, in conjunction with the extended computational time typically required by numerical methods, due to the need to assimilate large amounts of data and incorporate them in the initial conditions of the models. The challenge becomes especially prominent in the case of convective precipitations, characterized by high rainfall rates originating from cells spanning a few tens of kilometers \cite{Challenges2014}. 

Numerous historical datasets are accessible for short and medium-range weather forecasts, and they commonly exhibit a structured spatial grid format that bears a striking resemblance to images. Each weather variable can be represented as a map, or when considering the time dimension as a video.
The problem of predicting the next frame in a video sequence is a well-known image processing problem, and various categories of generative models have demonstrated their effectiveness in predicting the next frame in video sequences \cite{simvp,temporal_attention,msstnet2022,9063513}; it is not surprising that many recent studies focused on the use of deep neural network (DNN) architectures for weather nowcasting \cite{LSTM15_nowcasting,gmd-13-2631-2020,atmo_mdpi2020,metnet,TRU-NET,Deep12,bi2022panguweather,Hatanaka2023DiffusionMF}. All these models do not rely on explicit physical laws describing the dynamics of the atmosphere, adopting instead a backpropagation-based learning method to directly forecast the weather using observed data.

Most of the aforementioned models are trained to minimize the loglikelihood of the prediction, measured through a metric like mean squared error (MSE). As it is well known by other applications in image processing, loglikelihood-based optimization in the case of multimodal output typically results in averaging, introducing blurriness in the prediction.
As the lead time increases, the predicted fields become weaker and more widespread, suffering from the growing uncertainty in weather predictability.

Instead of predicting the expected amount of precipitation, it is possible to address its probability distribution, which is precisely the purpose of generative modeling \cite{generative_survey_2018,generative_introduction}. Usually, the probability distribution is learned in an implicit way, in the form of a {\em generator} able to sample data in accordance with the given distribution. 

For several years, the most representative class of generative models has been that of Generative Adversarial Networks (GANs) \cite{GANs,gan_survey}.
In the case of GANs, the training process involves a generator, which acts as a sampler for the intended distribution, and a discriminator, responsible for evaluating the generator's output by discerning between real and generated (``fake'') data. This training can be conceptualized as a zero-sum game, where the gain of one agent corresponds to the loss of the other. The generator and discriminator are alternately trained, with each adversarial component being frozen during the training of the other. Ultimately, the objective is for the generator to prevail, generating samples that the discriminator is unable to differentiate from real data. GANs typically offer better generative quality than likelihood-based models like Variational Autoencoders \cite{RezendeMW14,VAEKingma,VAEGreen}, possibly at the price of a reduced sampling diversity \cite{comparingNCAA}. They are, in fact, prone to {\em mode collapse} \cite{mode-collapse}, in which the generator learns to output just one or a few different examples, for instance, ignoring its noise input, and therefore generating identical outputs for a given input.

In the field of weather forecasting, GANs have been used for downscaling \cite{LeinonenNB21,Price22,GAN_downscaling22}, precipitation estimation from remote satellite sensors \cite{satellite2019,PrecipGan}, and disaggregation \cite{disaggregation-2021}. The Deep Generative Models of Rainfall (DGMR)  \cite{nature_nowcasting} is usually reputed to be the best generative nowcasting model based on GANs: it uses a conditional GAN with a regularization term to incentivize the model to produce forecasts close to the true precipitation. 

\begin{figure*}[t]
    \centering
    \includegraphics[width=\textwidth]{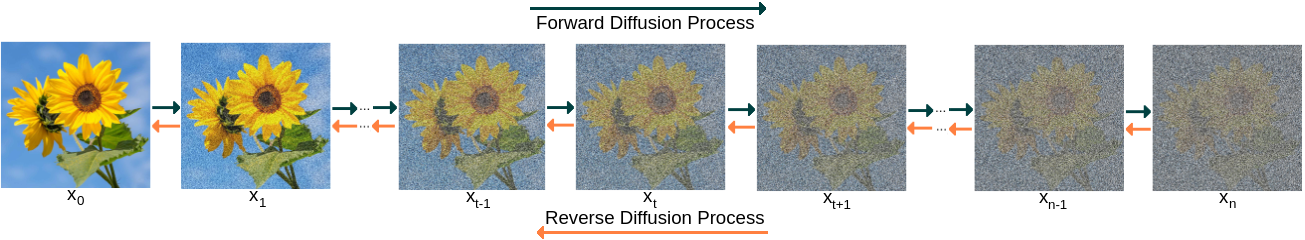}
    \caption{Forward (from left to right) and reverse (from right to left) diffusion process.}
    \label{fig:reverse-diffusion}
\end{figure*}

The leading role traditionally held by GANs in the field of generative modeling has been recently challenged by Denoising Diffusion Models (DDM) \cite{DDPM}. 
This generative technique possesses distinctive characteristics that have been crucially exploited in many recent and well-known applications \cite{DALLE2,Imagen}, comprising video generation \cite{imagen_video, video_diffusion2022}. These attributes include exceptional generation quality, a strong sensitivity and responsiveness to conditioning, diverse sampling capabilities, training stability, and satisfactory scalability \cite{Diff_vs_GAN, asperti2022embedding}. 

In essence, a diffusion model trains a single network to eliminate noise from images, where the level of noise to remove can be parametrically set. The network is then used to generate new samples by progressively reducing, in an iterative loop, the amount of noise in a given ``noisy'' image, starting from a completely random noise configuration. Traditionally referred to as reverse diffusion, this process aims to ``invert'' the direct diffusion process, which consists in iteratively adding noise to the source image (see Figure \ref{fig:reverse-diffusion}).

The idea of applying diffusion models for weather forecasting is quite natural, and several teams are independently working around this problem at the moment, on different datasets. In \cite{harris2022generative} diffusion models have been successfully applied to a downscaling problem. In \cite{Leinonen23}, a diffusion model has been trained on a dataset of the MeteoSwiss operational radar network \cite{suiss_dataset_2006,willemse2016weather}, and tested over 
data obtained from the radar composite of the German Weather Service (DWD) \cite{German_2008} from April–September 2022. Results have been compared with a GAN-based Deep Generative Models of Rainfall (DGMR) and a statistical model, PySTEPS, showing a sensible improvement both in accuracy and diversification. A detailed comparison of this architecture with our model is given in Section~\ref{sec:difference}.

In this article, we compare a different diffusion model with the Weather Fusion UNet (WF-UNet) model in \cite{WF-UNet}, integrating precipitation and wind speed variables as input of the learning process. Data refer to six years of precipitation and wind radar images from Jan 2016 to Dec 2021 of 14 European countries, with 1-hour temporal resolution and 31 square km spatial resolution based on the traditional ERA5 dataset \cite{https://doi.org/10.1002/qj.3803}. 
On the typical Mean Squared Error metric, our diffusion model outperforms WF-UNet. 

The article follows the subsequent structure. It begins by addressing theoretical aspects related to diffusion models and relative conditioning. Subsequently, it introduces the dataset and experimental setting. The methodology is then presented, showcasing the novel Generative Ensemble Diffusion (GED) approach, and comparing it with related models. The article further discusses the conducted experiments and concludes by delving into the implications of the research.

\section{Diffusion Models}\label{sec:diffusion_theory}

Diffusion models are a class of probabilistic generative models that are particularly effective in modeling complex, high-dimensional data distributions and have found applications in various domains such as computer vision, natural language processing, and generative art. At the core of diffusion models lies the mathematical concept of a diffusion process, that is, a stochastic process that describes the continuous random movement of particles over time, modeling the spread or diffusion of some quantity in space or time, where the particles tend to move from regions of high concentration to regions of low concentration, resulting in a gradual blending or mixing of the quantity. In the context of machine learning, diffusion models leverage the principles of diffusion processes to model the generation of data. Instead of directly sampling data points from a fixed distribution, these models iteratively transform a simple initial distribution, typically a known distribution like a Gaussian or uniform distribution, into the desired complex data distribution. The main idea is to perform a series of diffusion steps, where each step updates the probability distribution of the data. This is achieved by adding Gaussian noise to the current data samples and iteratively refining them. 

From a mathematical perspective, considering a distribution $q(x_0)$ which generates the data, generative models aim to find a parameter vector $\theta$ such that the distribution $p_\theta(x_0)$ parameterized by a neural network approximates $q(x_0)$. 

Denoising Diffusion Probabilistic Models (DDPM)\cite{DDPM} assume the generative distribution $p_\theta(x_0)$ to have the form
\begin{equation}
p_\theta(x_0))\int p_\theta(x_{0:T})dx_{1:T}
\end{equation}
given a time range horizon $T>0$.
where
\begin{equation}
p_\theta(x_{0:T})=p_\theta(x_T)\prod_{t=1}^{T}p_\theta(x_{t-1}\vert x_t).
\end{equation}
%with 
%\begin{equation}
%p_\theta(x_T)=\mathcal{N}(x_T\vert 0; I)
%\end{equation} 
%and 
%\begin{equation}
%p_\theta(x_{t-1} \vert x_t))\mathcal{N}(x_{t1}\vert %\mu_\theta(x_t,\alpha_t);\omega_t^2I).
%\end{equation}

Training is traditionally based on a variational lower bound of the negative loglikelihood:
\begin{align}
& - \log p_\theta(x_0)  \nonumber \\ 
& \hspace{.4cm}\leq - \log p_\theta(x_0) \!+\! D_\text{KL}(q(x_{1:T}\vert x_0) \| p_\theta(x_{1:T}\vert x_0) ) \nonumber 
 \\
%\hspace{.4cm}= -\log p_\theta(x_0) + \mathbb{E}_{x_{1:T}\sim q(x_{1:T} \vert x_0)} \Big[ \log\frac{q(x_{1:T}\vert x_0)}{p_\theta(x_{0:T}) / p_\theta(x_0)} \Big] \\
& \hspace{.4cm}= -\log p_\theta(x_0) \! +\! \mathbb{E}_q \Big[ \log\frac{q(x_{1:T}\vert x_0)}{p_\theta(x_{0:T}) / p_\theta(x_0)} \Big] \nonumber \\
& \hspace{.4cm}= -\log p_\theta(x_0) \!+\! \mathbb{E}_q \Big[ \log\frac{q(x_{1:T}\vert x_0)}{p_\theta(x_{0:T})} + \log p_\theta(x_0) \Big] \nonumber \\
& \hspace{.4cm}= \mathbb{E}_q \Big[ \log q(x_{1:T}\vert x_0) -p_\theta(x_{0:T}) \Big] = \mathcal{L}(\theta) \label{eq:loss_elbo}
\end{align}

Differently from typical latent variable models like Variational Autoencoders (VAEs) \cite{RezendeMW14,VAEKingma,VAEGreen}, diffusion models employ a fixed (non-trainable) inference procedure 
$q(x_{1:T} \vert x_0)$. Additionally, latent variables are characterized by relatively high dimensionality, usually identical to the dimensions of the visible space.

In the particular case of Denoising Diffusion Implicit Models (DDIMs)\cite{2020arXiv201002502S}, used
in this work, the authors considered a non-Markovian diffusion process
\begin{equation}
    q_\sigma (x_{1:T} \vert x_0) = q_\sigma(x_T \vert x_0) \prod_{t=2}^T q_\sigma (x_{t-1} \vert x_t, x_0)
\end{equation}
where $q_\sigma(x_T \vert x_0) = \mathcal{N}(x_T \vert \sqrt{\alpha_T} x_0, (1 - \alpha_T) \cdot I)$, and

\begin{align}\label{eq:non_markovian_reverse_diffusion}
    q_\sigma (x_{t-1} \vert x_t, x_0) = \mathcal{N} \Bigl( x_{t-1} \Big \vert \mu_{\sigma_t}(x_0, \alpha_{t-1}); \sigma_t^2 \cdot I \Bigr)
\end{align}
with 
\[
\begin{array}{l}
    \mu_{\sigma_t}(x_0, \alpha_{t-1}) = \\
    \hspace{.5cm}\sqrt{\alpha_{t-1}} x_0 + \sqrt{1 - \alpha_{t-1} - \sigma_t^2} \cdot \frac{x_t - \sqrt{\alpha_t} x_0}{\sqrt{1 - \alpha_t}}.
\end{array}
\]
The definition of $q(x_{t-1} \vert x_t,x_0)$ has been cleverly chosen to respect two important aspects of the diffusion process of DDPM: the Gaussian nature of $q(x_{t-1} \vert x_t,x_0)$ (once conditioned on $x_0$) and the fact that the marginal distribution $q_\sigma(x_t\vert x_0) = \mathcal{N}(x_t \vert \sqrt{\alpha_t} x_0; (1 - \alpha_t) \cdot I)$, recovers the same marginals as in DDPM. As a consequence of the latter property, we can express $x_t$ 
as a linear combination of $x_0$ and a noise variable $\epsilon_t \sim \mathcal{N}(\epsilon_t \vert 0; I)$:
\begin{align}\label{eq:xt_from_x0}
    x_t = \sqrt{\alpha_t} x_0 + \sqrt{1 - \alpha_t} \epsilon_t.
\end{align}

Next, we need to define a trainable generative process $p_\theta(x_{0:T})$ where  $p_\theta(x_{t-1}\vert x_t) $ leverages the structure of $q_\sigma(x_{t-1} \vert x_t, x_0)$. The idea is that given a noisy observation $x_t$, one starts making a prediction
of $x_0$, and then use it to obtain $x_{t-1}$ 
according to equation~\ref{eq:non_markovian_reverse_diffusion}, 
that is.

In practice, we train a neural network $\epsilon_\theta^{(t)}(x_t, \alpha_t)$ to map a given $x_t$ and a noise rate $\alpha_t$ to an estimate of the noise $\epsilon_t$ added to $x_0$ to construct $x_{t}$. Consequently, $p_\theta(x_{t-1} \vert x_t)$ becomes a $\delta_{f_\theta^{(t)}}$, where
\begin{align}\label{eq:approx_DDIM}
    f_\theta^{(t)}(x_t, \alpha_t) = \frac{x_t - \sqrt{1 - \alpha_t} \epsilon_\theta(x_t, \alpha_t)}{\sqrt{\alpha_t}}.
\end{align}
Using $f_\theta^{(t)}(x_t, \alpha_t)$ as an approximation of 
$x_0$ at timestep $t$, $x_{t-1}$ is then obtained as follows:
\begin{align}
x_{t-1} = & \sqrt{\alpha_{t-1}}\cdot f_\theta^{(t)}(x_t, \alpha_t) + \nonumber \\ 
& 
\sqrt{1-\alpha_{t-1}-\sigma_t^2}\cdot \epsilon_\theta(x_t, \alpha_t) \label{eq:x_t_minus_1}
\end{align}.

As for the loss function, the term in eq.\ref{eq:loss_elbo}
can be further refined expressing $L_\theta$ as the sum of the following terms \cite{Sohl-DicksteinW15}:
\begin{equation} 
\label{eq:loss_sum}
L_\theta = L_T + L_{t-1} + \dots + L_0 
\end{equation}
where
\[
\begin{array}{l}
L_T = D_\text{KL}(q(x_T \vert x_0) \parallel p_\theta(x_T)) \\
L_t = D_\text{KL}(q(x_t \vert x_{t+1}, x_0) \| p_\theta(x_t \vert x_{t+1})) \\
\hspace{3.5cm}\text{ for }1 \leq t \leq T-1 \\
L_0 = - \log p_\theta(x_0 \vert x_1)
\end{array}
\]
All previous distributions are Gaussians and their KL divergences 
can be calculated in closed form, in the Rao-Blackwellized fashion.
After a few manipulations, we get to the following formulation:
\begin{equation}
L_t = \mathbb{E}_{t \sim [1, T], x_0, \epsilon_t} \Big[\gamma_t\| \epsilon_t - \epsilon_\theta(x_t, t) \|^2 \Big]
\end{equation}
that can be interpreted as the weighted mean squared error between the predicted and the actual noise a time $t$. 

The weighting parameters are frequently ignored in practice, since 
experimentally the training process seems to work better without them. 

The pseudocode for training and sampling is given in the following Algorithms. %~\ref{algorithm1,algorithm2}.

\begin{algorithm}[H]
    \centering
    \caption{Training\label{algorithm1}}
    \begin{algorithmic}[1]
        \Repeat
        \State $x_0 \sim q(x_0) $  %\Comment{take a sample }
        \State $t \sim $Uniform({1,..,T}) %\Comment{choose a timestep}
        \State $\epsilon \sim \mathcal{N}(0;I)$ %\Comment{create random gaussian noise}
        \State $x_t = \sqrt{\alpha_t} x_b + \sqrt{1\!-\! \alpha_t} \epsilon$  %\Comment{corrupt the sample with noise rate $\alpha_t$}
        \State Backpropagate on
        $\lvert \lvert \epsilon - \epsilon_{\theta} (x_t, \alpha_t )\rvert \rvert^2$ %\Comment{backpropagate the loss}
        \Until converged
    \end{algorithmic}
\end{algorithm}

Sampling is an iterative process, starting from a purely noisy image $x_T \sim \mathcal{N}(0,I)$. The denoised version of the image at timestep $t$ is obtained using equation~\ref{eq:x_t_minus_1}.

\begin{algorithm}[H]
    \centering
    \caption{Sampling}\label{algorithm2}
    \begin{algorithmic}[1]
        \State $x_T \sim \mathcal{N}(0,I)$
        %\Comment{get a purely noisy image}
        \For {$t = T,...,1$}
       \State $\epsilon = \epsilon_\theta(x_a,x_t,\alpha_t)$ %\Comment{predict noise}
       \State $\tilde{x}_0 = \frac{1}{\sqrt{\alpha_t}} (x_t - \frac{1-\alpha_t}{\sqrt{1 - \alpha_t}} \epsilon)$ %\Comment{compute denoised result}
       \State $x_{t-1} = \sqrt{\alpha_{t-1}}\tilde{x}_0 + \sqrt{1 - \alpha_{t-1}}\epsilon$  %\Comment{re-inject noise at rate $\alpha_{t-1}$}
        \EndFor
    \end{algorithmic}
\end{algorithm}

\subsection{Conditioning}
Generation often requires a way to control how samples are created to 
influence the final output. This process is commonly referred to as \textit{conditioned} or \textit{guided} diffusion. Numerous approaches have been devised to integrate image and/or text embeddings into the diffusion process, allowing for guided generation. In the mathematical context, ``guidance'' entails conditioning a prior data distribution, represented by $p(x)$, with a particular constraint, like a class label or an image/text embedding. This conditioning leads to the formation of a conditional distribution, denoted as $p(x\vert y)$.

To convert a diffusion model $p_\theta$ into a conditional diffusion model, we can introduce conditioning information $y$ at each diffusion step as follows:
\begin{equation}
    p_\theta(x_{0:T}\vert y) 
    = p_\theta(x_{T}) \prod_{t=1}^T p_\theta (x_{t-1} \vert x_t , y)
\end{equation}  

There are typically two approaches to learning this distribution, one based on 
an auxiliary classifier (similar, in spirit, to AC-GANs \cite{AC-GANS}), and 
a second one that is classifier-free.

The idea behind classifier guidance is the following. Our aim is to learn the gradient of the logarithm of the conditional density $p_\theta (x_t \vert y)$. By applying Bayes' rule, we can express it as:
\begin{equation}
    \nabla_{x_t} \log p_\theta ( x_t \vert y)
    = \nabla_{x_t} \log \left( \frac{p_\theta(y \vert x_t) \cdot p_\theta(x_t) }{p_\theta(y)} \right)
\end{equation}   
Since the gradient operator only applies to $x_t$, we can eliminate the term $p_\theta(y)$; after simplification we get:  \begin{align} \label{diffusion_guidance_formula}
    \nabla_{x_t} \log p_\theta ( x_t \vert y)
    = &\nabla_{x_t} \log p_\theta (x_t) \nonumber \\
    & + s \cdot  \nabla_{x_t} \log p_\theta (y \vert x_t) 
\end{align}   
Here $s$ is a scalar term used to modulate the strength of the guidance term.

As described in \cite{Diff_vs_GAN}, we can use a classifier $f_\phi(y \vert x_t, t))$ to guide the diffusion during generation. This technique involves training a classifier $f_\phi(y \vert x_t, t)$ on a noisy image $x_t$ to predict its class $y$. The gradient $\nabla_x \log f_\phi(y \vert x_t)$ can then be utilized to guide the diffusion sampling process towards the conditioning information $y$ by modifying the noise prediction. We shall not use this technique, particularly suited for
discrete labels, so further details are omitted.

%                   In particular, in \cite{classifier_guidance_paper} is showed that:
%                    \begin{equation}
%                        \begin{aligned}
%                            \nabla_{x_t} \log p_\theta(x_t, y)
%                            &= \nabla_{x_t} \log p_\theta(x_t) + \nabla_{x_t} \log p_\theta(y \vert x_t) \\
%                            &= - \frac{1}{\sqrt{1 - \bar{\alpha}_t}} (\boldsymbol{\epsilon}_\theta(x_t, t) - \sqrt{1 - \bar{\alpha}_t} \nabla_{x_t} \log f_\phi(y \vert x_t))
%                        \end{aligned}
%                    \end{equation} 
%                    Therefore, a new classifier-guided predictor $\hat{\boldsymbol{\epsilon}}_\theta$ can be expressed as:
%                    \begin{equation}
%                        \hat{\boldsymbol{\epsilon}}_\theta(x_t, t) = \boldsymbol{\epsilon}_\theta(x_t, t) - \sqrt{1 - \bar{\alpha}_t} \; \cdot s \cdot \nabla_{x_t} \log f_\phi(y \vert x_t)
%                    \end{equation} 
%                    As before, $s$ is a scalar term that modulates the strength of the guidance term.
                
The theory of condition diffusion without relying on an independent classifier has been investigated in \cite{classifier_free_diffusion}. The approach consists in training a conditional diffusion model $\epsilon_\theta(x_t, t, y)$ along with an unconditional model $\epsilon_{\theta}(x_t, t, 0)$. Typically, the same neural network can be used for both models: during training, the class $y$ is randomly set to 0, exposing the model to both conditional and unconditional setups. The estimated noise $\hat{\epsilon}_\theta(x_t \vert t, y)$ at timestep $t$ is then a suitably weighted combination  of the conditional and unconditional predictions:                    
\begin{equation}                         
    \hat{\epsilon}_\theta(x_t, t, y) = \epsilon_\theta(x_t, t, y) + s \cdot \epsilon_\theta(x_t, t)
\end{equation}

\begin{figure*}[h]
    \centering
    \includegraphics[width=0.92\textwidth]{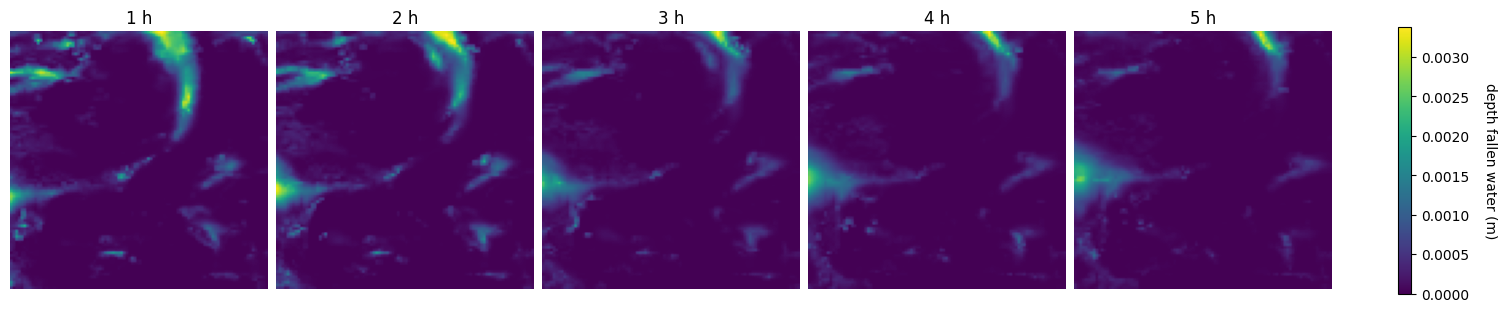}
    \caption{Example of precipitation data from the ERA-5 dataset}
    \label{fig:dataset-rain}
\end{figure*}

\section{Dataset description and preprocessing}
To assess the performance of our model we recreate the same samples as proposed in \cite{WF-UNet,BroadUnet,smatUnet}, focusing on the task of precipitation nowcasting. We selected for our research a subset of the ERA-5 dataset \cite{https://doi.org/10.1002/qj.3803}, which is a state-of-the-art global atmospheric reanalysis produced by the European Centre for Medium-Range Weather Forecasts (ECMWF). ERA-5 reports a comprehensive numerical representation of the Earth's recent climate history, spanning several decades and covering the entire globe at a high spatial resolution of approximately 31 kilometers. It provides hourly estimates of a multitude of atmospheric, land, and oceanic climate variables, such as temperature, precipitation, humidity, wind speed and direction, and sea surface temperature, among others. The ERA-5 dataset is the product of an advanced and uniform data assimilation system, melding millions of disparate observations with intricate Earth system modeling. This integration yields a coherent and consistent dataset, highly regarded and extensively used across various fields. Its applications span weather forecasting, climate studies, hydrological research, energy production prediction, and numerous other scientific domains, as well as policy-related endeavors.

With our task being precipitation nowcasting, our main target feature is the Total Precipitation variable, described as the accumulated liquid and frozen water that falls to the Earth’s surface, comprising of rain and snow. Our selected region of interest is defined by a geographical rectangle, with its latitudinal boundaries extending from latitude -12$^\circ$  to latitude 12 $^\circ$  and its longitudinal boundaries extending from longitude 36$^\circ$ to longitude 60$^\circ$. Images from this region cover much of the western part of Europe, partially covering 14 countries. Our collected time span covers a six-year period, from 2016 to 2021, with hourly measurements. The collected data have a dimension of 96×96 values with each representing the depth fallen water would have if it were spread evenly over the grid box of 31 square kilometers. The units of this parameter are depth in meters of water equivalent.

The dataset is normalized by dividing the values of both the training and testing set by the highest occurring value in the training set, we then split the dataset into a training set (years 2016-2020) and a testing set (year 2021). 

Rain is a sparse parameter, and it is often non-present in the area of analysis. This produces a dataset with an extensive amount of noninformative data which can bias the model towards a zero prediction \cite{smatUnet}. Therefore, we defined the generator for our data with an additional parameter, so that it returns only sequences with at least a percentage amount of rain in the pixels, therefore simulating the EU-50 and EU-20 datasets specified in \cite{WF-UNet}, whose images have at least 50\% and 20\% of rain in the pixels respectively. This selection operation is performed by computing the number of non-zero pixels on a wider region of size 105x173, the image is then cropped to the final dimension of 96x96.

\subsection{Additional features}
In precipitation nowcasting it is often necessary for the predictive models to consider a range of meteorological features beyond the presence or absence of rain. Variables such as temperature, pressure, humidity, wind direction, and wind speed can all significantly influence precipitation patterns. These factors, among others, interact in complex ways to shape the dynamics of the atmospheric system.
Another topic of concern is the level of awareness the model has of the underlying physical structure, and elements such as time embeddings, land/sea mask, and elevation information may be valuable for enabling the model to make a more informed decision.
In particular, in our work we experimented with wind speed, obtained from the two different northerly and easterly wind components, the land-sea mask, a geopotential map, and a sinusoidal time embedding, as reported in Table~\ref{tab: additional_info}.
All obtained from the ERA-5 dataset and normalized between 0 and 1 before training: 

\begin{table*}[h]
\centering
\def\arraystretch{1}% for the vertical padding
\begin{tabular}{>{\bfseries}lcp{8cm}}
\toprule
%\multicolumn{3}{c}{\textbf{Additional informations}} \\
\textbf{Name} & \textbf{Unit} & \textbf{Description}\\ 
\midrule
\midrule

100m wind speed & $ms^{-1}$ & Wind speed of air at a height of 100 meters above the surface of the Earth, given easterly and northerly components $u$ and $v$ the speed is obtained by $\sqrt{(u^2 + v^2)}$ \\ 
Timestamp & [m,d,h] & timestamp including month, day, and hour of the start of the given sequence, tile encoded into a 96x96 array. \\ 
Land-sea mask & dimensionless & Proportion of land, as opposed to ocean or inland waters in a grid box \\ 
Geopotential & $m^{2}s^{-2}$ & Gravitational potential energy of a unit mass, at a particular location at the surface of the Earth, relative to mean sea level. \\ 

\bottomrule
\end{tabular}
\caption{Additional features units and details}
\label{tab: additional_info}
\end{table*}

% \begin{itemize}
%     \item 10m u-component of wind, it is the horizontal speed of air moving towards the east, at a height of ten metres above the surface of the Earth, in metres per second.
%     \item 10m v-component of wind, it is the horizontal speed of air moving towards north, at a height of ten metres above the surface of the Earth, in metres per second.
%     \item Land-Sea mask, binary value representing the presence of land or sea. This parameters does not vary in time.
%     \item Geopotential, This parameter is the gravitational potential energy of a unit mass, at a particular location at the surface of the Earth, relative to mean sea level. This parameter does not vary in time. 
% \end{itemize}

\begin{figure}[h]
    \centering
    \includegraphics[width=0.5\textwidth]{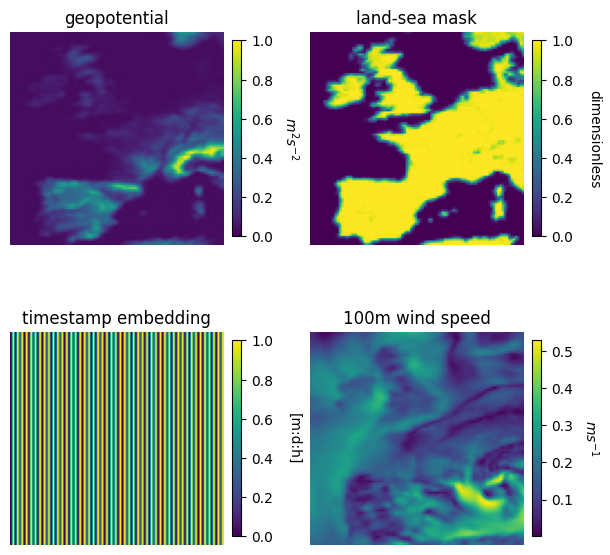}
    \caption{Visual example of the additional features}
    \label{fig:dataset-rain}
\end{figure}

\section{Methodology}
In this section, we introduce our proposed model, Generative Diffusion Ensemble (GDE). Our model is based on the Denoising Diffusion Implicit Model (DDIM), which is trained on a sequence of rain data augmented with other meteorological features. The model generates multiple outputs that are then synthesized into a final prediction through a U-Net architecture. This combination provides a comprehensive precipitation forecast, leveraging both the capability of DDIM to model the probability distribution of the weather data and the feature extraction strength of the U-Net.

We start by introducing the distinct components that form our GDE model, offering insight into their specific roles within the broader system. The discussion ultimately converges on the overall structure of the model.

\subsection{The DDIM architecture}

Diffusion models essentially operate as iterative denoising algorithms. Their main trainable component is the denoising network, denoted as $\epsilon_\theta(x_t, \alpha_t)$. This network receives as input the noisy images, $x_t$, and a corresponding noise variance, $\alpha_t$, with the objective of estimating the amount of noise infiltrating the image. The training of this underlying denoising network is done conventionally. An initial sample, $x_0$, is extracted from the dataset and subjected to a predefined amount of random noise. The network is then tasked with estimating the noise within these corrupted images.

\subsubsection{denoising}

Our model of choice as the denoising network is a U-net. The U-net is one of the most common architectures for denoising \cite{9360532,LEE202092,heinrich2018residual,komatsu2020comparing} and it is often implemented in diffusion models \cite{dhariwal2021diffusion}. Originally introduced for semantic segmentation \cite{U-net}, the U-Net architecture has gained widespread popularity and found applications in diverse image manipulation tasks. The network comprises a downsampling sequence of layers, followed by an upsampling sequence while incorporating skip connections between layers of the same size. Typically, the U-Net's configuration is determined by defining the number of downsampling blocks and the number of channels for each block. The upsampling structure follows a symmetric pattern and the spatial dimension is dependent on the image resolution, which in our case is 96x96. 
Consequently, a U-Net's entire structure can be encoded concisely in a single list, such as [32, 64, 96, 128]. This list represents both the number of downsampling blocks (in this case, 4) and the corresponding number of channels, which usually increase as the spatial dimension decreases. For our experiments we selected a U-net size of [64, 128, 256, 384] which proved to be the most effective experimentally. To improve the sensibility of the U-net to the noise variance, $\alpha_t$ is taken as input, which is then embedded using an ad-hoc sinusoidal transformation by splitting the value in a set of frequencies, in a way similar to positional encodings in Transformers \cite{attention}. The embedded noise variance is then vectorized and concatenated to the noisy images along the channel axes before being passed to the U-Net. This helps the network to be highly sensitive to the noise level, which is crucial for good performance. We implement sinusoidal embeddings using a Lambda layer.

\subsubsection{conditioning}

Conditioning of the model is necessary to guide the diffusion towards a forecast defined by the known previous weather conditions. Our conditioning is applied in a classifier-free manner, by concatenating the conditioning frames to the noisy images alongside the channel axis. 

Practically, the model $\epsilon_\theta(x_t, t, y)$ takes as input the noisy images $
x_t = \{r_1, r_2 , r_3\}$ where $r_h$ represents the future prediction of the rain precipitation $h$ hours ahead. The conditioning information $y = \{r_{-8..0},u_{-1,0},v_{-1,0},lsm,geopot\}$  contains the previous 8 hours of precipitation information $r_{-11..0}$, the previous two hours for both wind components $u_{-1,0},v_{-1,0}$ and two static maps representing the land-sea mask $lsm$ and the geopotential $geopot$.

\begin{figure*}[h]
    \centering
    \includegraphics[width=0.7\textwidth]{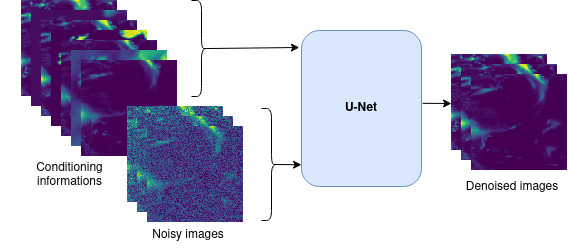}
    \caption{Conditioning is implemented by stacking additional information alongside the channel axis in the denoising network}
    \label{fig:dataset-rain}
\end{figure*}

Our implementation directly provides the U-net with the conditioning information, specifically, each temporal slice in the input data is treated analogously to a color channel in an RGB image. By applying 2D convolutions across these temporal slices independently our model is able to extract sufficient frame-level temporal features to effectively produce a sequence coherent with the past frames used as conditioning. 

For example, when training with a batch size of 16, our input data to the denoising network would have the shape of [16,96,96,17], with the last dimension containing both the conditioning information and noisy images. Our output on the other hand would comprise only the denoised 3 frames, therefore would have the shape of [16,96,96,3].

Interestingly, our training process was proven successful in consistently achieving temporal conditioning training exclusively with conditioned instances, whereas examples in the literature required alternating between conditioned and non-conditioned training instances, followed by a weight mixing stage \cite{ho2022classifier}.

\subsection{The proposed Generative Diffusion Ensemble (GED)}
Our proposed Generative Diffusion Ensemble (GED) approach leverages the power of diffusion to integrate the inherent probability distribution of meteorological patterns, which is then used to synthesize a probable precipitation prediction. Image samples generated through the diffusion process, in line with the generative traits of the model, yield a highly diverse set of outputs despite sharing identical conditioning information. Based on the assumption that the diffusion model captures the stochastic essence of weather dynamics, a prediction can be derived from an ensemble of possible outcomes. 

The methodology we propose aligns fundamentally with the ensemble post-processing techniques conventionally utilized in Numerical Weather Prediction (NWP) models. Ensemble post-processing in weather forecasting refers to the statistical refinement of a set of multiple weather forecasts produced from slightly different initial conditions \cite{vannitsem2018statistical}. These multiple simulations provide a range of possible weather outcomes and function as an estimate of forecast uncertainty.
The methods employed in ensemble post-processing exhibit considerable diversity and range from statistical techniques such as linear regression and distributional regression \cite{gneiting2005calibrated,henzi2021isotonic}, to more sophisticated machine learning algorithms like QRF \cite{taillardat2016calibrated} or EMOS-GB \cite{messner2017nonhomogeneous}. In recent years most of the research interest focused on Neural methods for ensemble post-processing, with notable successful examples in literature.  \cite{ENS-10,MachineLearningEnsembleComparison}.

In our work, we propose two different approaches to executing the ensemble prediction, one based on a simple statistical method and the second utilizing a neural model. The first approach involves executing multiple diffusion generations in an iterative manner (parallelizable across the batch dimension with a significant speedup \cite{alpha-flops}) and subsequently calculating the mean of these generated images. This method effectively condenses the probability distribution of the images into an average outcome, thereby yielding a more accurate forecast. Our experimental results have demonstrated that this strategy of computing the mean of multiple generations produces superior results when compared to utilizing a single diffusion generation. Thus, this approach effectively leverages the multitude of potential outcomes to generate a more accurate and robust prediction.

\begin{figure*}[h]
    \centering
    \includegraphics[width=\textwidth]{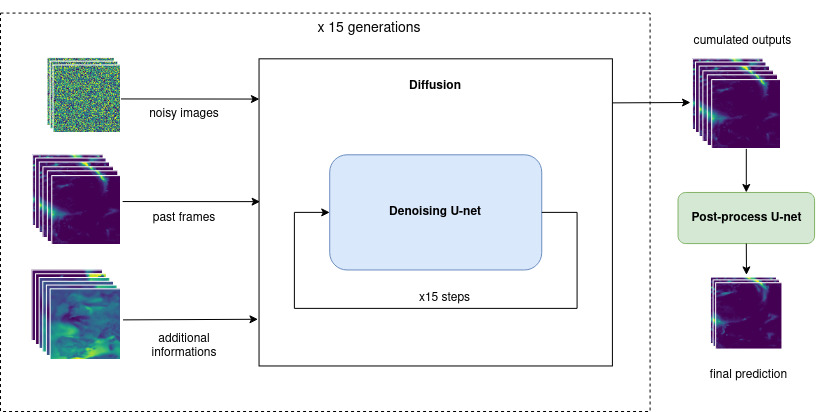}
    \caption{Generative Diffusion Ensemble (GED) prediction structure, showing the multiple denoising cycles and the final post-processing step.}
    \label{fig:dataset-rain}
\end{figure*}

In our second approach, instead of simply computing a mean, we delegate the synthesis of the prediction to a more sophisticated module, similar to what is frequently done in the literature \cite{steps,steps2}. 
In our case, we use a U-Net architecture to amalgamate the generated outcomes into a more probable prediction. 
This strategy not only allows for a more informed decision-making process in integrating the outputs but also provides the opportunity to add a post-processing layer specifically trained on our target image, in contrast to the diffusion model which is trained on the noise difference of the single diffusion steps. 
%the Mean Absolute Error (MAE) pertaining only to noise. which is the Mean Squared Error (MSE) on the output image, in contrast to the diffusion model trained on the Mean Absolute Error (MAE) pertaining only to noise. (debole: perchè non usare mse anche sulla diffusion allora?) (vero, però non è proprio la stessa cosa, la diffusion non applica la loss sull'immagine finale, ma sui singoli passaggi di denoise. Provo a cambiare così.) 
Our experimental findings have validated the efficacy of this methodology, demonstrating it to consistently produce superior results to both simple diffusion and diffusion ensemble with the mean method.

\subsection{Training and Evaluation}
Our diffusion model underwent training with a batch size of 2 over 40 epochs, using the AdamW optimization algorithm \cite{loshchilov2017decoupled} with a learning rate of 1e-04 and weight decay of 1e-05. Interestingly, we observed that augmenting the batch size hampered the training process. Furthermore, a fine-tuning phase of 10 epochs using a learning rate of 1e-05 and weight decay of 1e-06 yielded marginal improvements in the overall results. Training was implemented via the use of a generator, which produced a random batch of sequences spanning the training years from 2016 to 2020. Mirroring our reference model, we omitted sequences composed of over 50\% non-rain values from the training process. Consistent with standard diffusion model implementations, our loss function was the Mean Absolute Error (MAE) applied to the difference in noise. Training experiments based on image loss, as opposed to noise, led to inferior results.

Our diffusion model underwent evaluation based on data from the test year of 2021. The number of diffusion steps was fixed at 15, with performance assessed using the Mean Squared Error (MSE) metric, defined in Equation~\ref{eq:mse}. Here, $n$ represents the total number of samples, $y_i$ denotes the ground truth value, and $\hat{y}_i$ signifies the predicted value. Please note that all metrics were computed on data post-denormalization. All training was conducted on an Nvidia RTX 4000 graphic card using the TensorFlow/Keras framework. 

\begin{equation}
MSE = \frac{1}{n} \sum_{i=1}^{n} (y_i - \hat{y}_i)^2
\label{eq:mse}
\end{equation}

The Generative Diffusion Ensemble (GDE) incorporates a post-processing U-net that shares the spatial dimensions of the denoising U-net, excluding the embedded variance. This network accepts fifteen distinct generative outputs from the diffusion model as input, each consisting of three future predictions for the subsequent three hours. The U-net then yields an output comprising three images, each predicting the rainfall for one of the upcoming three hours.

Training of this model parallels the process used for our diffusion model, employing AdamW as the optimizer with a learning rate of 1e-4 and a weight decay of 1e-5. The loss is computed using the Mean Squared Error (MSE) between the predicted images and their corresponding ground truth. For training, data is dynamically generated by the diffusion model using random sequences from the training years. Likewise, evaluation is performed on sequences from the test year of 2021.

\subsection{Differences with related models}\label{sec:difference}

To the best of our knowledge, the only other work currently addressing precipitation nowcasting using diffusion models is \cite{2023arXiv230412891L}. Comparing the two architectures is quite challenging due to the significantly different spatiotemporal scales of the data involved. In \cite{2023arXiv230412891L}, they consider time steps of 5 minutes, utilizing 4 time steps (20 minutes) of precipitation as input, and predict precipitation up to 20 time steps (100 minutes) into the future. Moreover, the geographical scale differs greatly as well, with radar signals collected at a 1 km resolution in a rectangular area spanning 710 km in the east–west direction and 640 km in the north–south direction, covering all of Switzerland and some surrounding regions.

In the cited work, the network is trained on 256x256 pixel images, corresponding to a geographical area of 256 $Km^2$. To manage computational costs, they adopt the notion of Latent Diffusion Model (LDM) popularized by Stable Diffusion \cite{stable_diffusion}, where the diffusion process runs in a latent variable space mapped to the physical pixel space through an autoencoder.

The diffusion model used in their work is quite different from ours, starting with the number of denoising iterations, which is 50 compared to our 15. Additionally, their Denoiser makes use of a forecaster stack based on Adaptive Fourier Neural Operators (AFNOs) \cite{2021arXiv211113587G}\cite{2022arXiv220211214P} to condition the model. They incorporate temporal cross-attention to map between the input and output time coordinates and a different denoiser stack that is also based on AFNOs, simulating cross-attention. The actual relevance of these modules is not documented, since no ablation was performed. We tested some of these solutions without noticing sensible improvements.

Other related works in precipitation nowcasting based on CNNs often utilize 3D convolutions or other conditioning methods such as RNNs or LSTMs to explicitly model temporal connections \cite{WF-UNet,smatUnet,BroadUnet,2020arXiv201103303G}. In recent years multiple publications have shown promising results in treating timesteps as multiple channels in the network, in this way achieving temporal prediction with only 2D convoluted layers \cite{AYZEL2019186,49235}.  In our proposed diffusion model, we handle temporal data using 2D convolutions in a similar manner to how color channels are dealt with in image processing. Specifically, each temporal slice in the input data is treated analogously to a color channel in an RGB image. By applying 2D convolutions across these temporal slices independently, akin to processing different color channels, our model was able to match the performance of competing 3D CNNs models.

\begin{table*}[h]
\centering
\def\arraystretch{1}% for the vertical padding
\begin{tabular}{>{\bfseries}lcccc}
\toprule
\multicolumn{5}{c}{\textbf{MSE values and additional metrics for EU20 dataset}} \\
\midrule
\textbf{Model} & \textbf{MSE} & \textbf{Accuracy} & \textbf{Precision} & \textbf{Recall} \\ 
\midrule
\multicolumn{5}{c}{1 hour ahead} \\
\midrule
Core U-net         & 2.97e-04 & 0.863 & 0.698 & 0.837 \\ 
Broad U-net        & 3.05e-04 & 0.861 & 0.706 & 0.803 \\ 
WF-UNet            & 2.67e-04 & 0.933 & 0.790 & 0.847 \\ 
Single Diffusion   & 2.86e-04 & 0.911 & 0.754 & 0.888 \\ 
GED (mean)         & 2.25e-04 & \underline{0.930} & 0.786 & 0.901 \\ 
GED (postprocess)  & \underline{2.03e-04} & 0.923 & \underline{0.798} & \underline{0.909} \\ 

\midrule
\multicolumn{5}{c}{2 hour ahead} \\
\midrule
Core U-net         & 5.02e-04 & 0.813 & 0.609 & 0.796 \\ 
Broad U-net        & 5.05e-04 & 0.819 & 0.638 & 0.712 \\ 
WF-UNet            & 4.87e-04 & 0.895 & 0.664 & 0.807 \\ 
Single Diffusion   & 4.69e-04 & 0.886 & 0.705 & 0.831 \\ 
GED (mean)         & 3.93e-04 & \underline{0.900} & 0.731 & 0.848 \\ 
GED (postprocess)  & \underline{3.53e-04} & 0.898 & \underline{0.742} & \underline{0.849} \\

\midrule
\multicolumn{5}{c}{3 hour ahead} \\
\midrule
Core U-net       & 6.71e-04 & 0.800 & 0.612 & 0.657 \\
Broad U-net      & 6.55e-04 & 0.806 & 0.637 & 0.609 \\ 
WF-UNet          & 6.34e-04 & 0.877 & 0.626 & 0.736 \\ 
Single Diffusion & 6.10e-04 & 0.853 & 0.638 & 0.758 \\ 
GED (mean)       & 5.20e-04 & 0.880 & 0.689 & \underline{0.801} \\ 
GED (postprocess)& \underline{4.70e-04} & \underline{0.891} & \underline{0.701} & 0.796 \\ 

\bottomrule
\end{tabular}
\caption{Results comparison on the EU20 Dataset}
\label{tab:20}
\end{table*}

\begin{table*}[htbp]
\centering
\def\arraystretch{1}% for the vertical padding
\begin{tabular}{>{\bfseries}lcccc}
\toprule
\multicolumn{5}{c}{\textbf{MSE values and additional metrics for EU50 dataset}} \\
\midrule
\textbf{Model} & \textbf{MSE} & \textbf{Accuracy} & \textbf{Precision} & \textbf{Recall} \\ 
\midrule
\multicolumn{5}{c}{1 hour ahead} \\
\midrule
Core U-net         & 3.18e-04 & 0.862 & 0.698 & 0.833 \\
Broad U-net        & 3.24e-04 & 0.860 & 0.705 & 0.795 \\ 
WF-UNet            & 2.50e-04 & 0.921 & 0.803 & 0.849 \\ 
Single Diffusion   & 2.59e-04 & 0.915 & 0.767 & 0.882 \\ 
GED (mean) & 2.02e-04 & \underline{0.924} & 0.782 & 0.885 \\ 
GED (postprocess) & \underline{1.99e-04} & 0.913 & \underline{0.803} & \underline{0.907} \\ 

\midrule
\multicolumn{5}{c}{2 hour ahead} \\
\midrule
Core U-net         & 5.02e-04 & 0.813 & 0.609 & 0.796 \\
Broad U-net        & 5.05e-04 & 0.819 & 0.638 & 0.712 \\ 
WF-UNet & 4.62e-04 & 0.877    & 0.684 & 0.813 \\ 
Single Diffusion   & 4.51e-04 & 0.875 & 0.699 & 0.844 \\ 
GED (mean)         & 3.59e-04 & \underline{0.882} & 0.711 & \underline{0.862} \\ 
GED (postprocess)  & \underline{3.40e-04} & 0.878 & \underline{0.724} & 0.860 \\

\midrule
\multicolumn{5}{c}{3 hour ahead} \\
\midrule
Core U-net       & 6.71e-04 & 0.800 & 0.612 & 0.657 \\
Broad U-net      & 6.55e-04 & 0.806 & 0.637 & 0.609 \\ 
WF-UNet          & 6.31e-04 & 0.855 & 0.647 & 0.743 \\ 
Single Diffusion & 6.03e-04 & 0.848 & 0.672 & 0.801 \\ 
GED (mean)       & 4.92e-04 & 0.856 & 0.701 & \underline{0.828} \\ 
GED (postprocess)& \underline{4.65e-04} & \underline{0.861} & \underline{0.706} & 0.821 \\ 

\bottomrule
\end{tabular}
\caption{Results comparison on the EU50 Dataset}
\label{tab:50}
\end{table*}

\begin{table*}[htbp]
\centering
\def\arraystretch{1}% for the vertical padding
\begin{tabular}{p{5cm}ccc}
\toprule
\multicolumn{4}{c}{\textbf{Single diffusion with different inputs on EU-50}} \\
\midrule
\textbf{Inputs} & \textbf{MSE 1h} & \textbf{MSE 2h} & \textbf{MSE 3h}  \\ 
\midrule
\midrule

8 rain & 2.62e-04 & 4.60e-04 & 6.21e-04  \\ 
8 rain + lsm + geopot & 2.60e-04 & 4.61e-04 & 6.23e-04  \\ 
8 rain + lsm + geopot + time & 2.60e-04 & 4.56e-04 & 6.16e-04  \\ 
8 rain + lsm + geopot + time + 2 wind speed & \underline{2.59e-04} & \underline{4.51e-04} & \underline{6.03e-04}  \\ 

\bottomrule
\end{tabular}
\caption{Results comparison on the EU20 Dataset}
\label{tab: inputs}
\end{table*}

\section{Experiments and Results}\label{sec:experiments}

In this section, we present our experimental setup and the conducted experiments.

All experiments were performed using models implemented in the TensorFlow/Keras framework. The training dataset consisted of precipitation data and additional features for the specified region, ranging from 2016 to 2021, while the test set exclusively utilized data collected in 2021. To compute the results, we performed an exhaustive analysis of all sequences within the given year for both the EU-20 and EU-50 datasets. 

All of our models compute the three different predictions simultaneously. It may be argued that with a unique model instance for each prediction, the overall performance may slightly improve, but this would happen at the expense of the training and inference time, with the latter being especially relevant for the operational application of precipitation forecasting. 

Our primary goal is to minimize the Mean Squared Error (MSE) within the initial three hours of the prediction model. Nonetheless, the determination of the optimal input data and model configuration remains a topic of ongoing debate. To address this, we conducted a series of initial tests aimed at identifying the most advantageous set of input features. In Table \ref{tab: inputs} we report a comparison of scores obtained with different sets of additional features using the Single Diffusion model.

\begin{figure}[h]
  \centering
  \begin{subfigure}[b]{0.49\textwidth}
    \includegraphics[width=\textwidth]{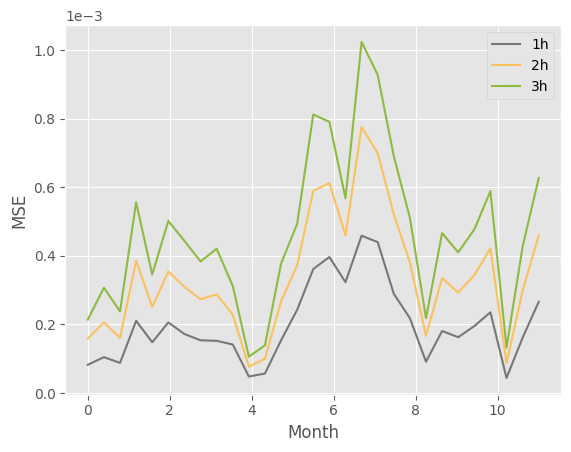}
    \caption{}
    \label{fig:image1}
  \end{subfigure}
  \hfill
  \begin{subfigure}[b]{0.49\textwidth}
    \includegraphics[width=\textwidth]{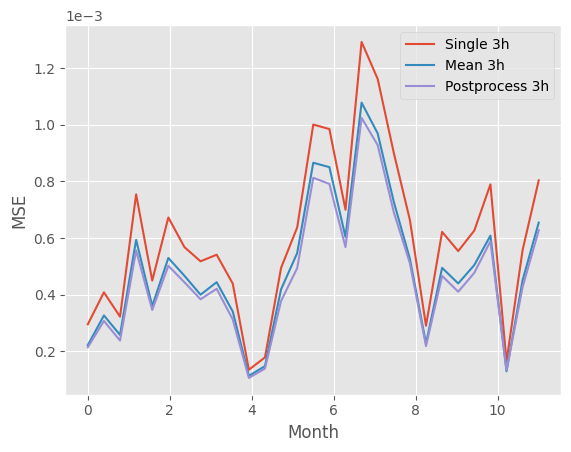}
    \caption{}
    \label{fig:image2}
  \end{subfigure}
  \caption{Single diffusion results for the year 2021 on EU50, showing high dissimilarity in score depending on the month of the year. In (a) we can see that the dissimilarity is present for each of the 3 predicted hours, in (b) note that the dissimilarity is not affected by computing the outcome with Single Diffusion, GDE (mean), or GDE (post-process).}
  \label{fig:figure}
\end{figure}

For what concerns model comparison, Table~\ref{tab:20} and \ref{tab:50} present a comparative analysis between a standard Core U-Net model, BroadU-Net \cite{BroadUnet}, WF-UNet (which incorporates additional features as proposed by \cite{WF-UNet}), our diffusion model with a singular generative output (Single Diffusion), and two distinct implementations of the Generative Ensemble Diffusion (GED). The GED models generate a final prediction by integrating 15 different generations of the three predicted frames. In the first implementation, the prediction is calculated by averaging all the values (mean), while the second version employs a post-processing U-Net for this task (post-process). The primary metric utilized for these results is Mean Squared Error (MSE), supplemented with the additional metrics of Accuracy, Precision, and Recall. Our findings suggest that, although a single diffusion prediction is outperformed by the U-Net models, both implementations of GED significantly surpass the performance of any U-Net-based approach. Among these, the GED version with post-processing demonstrated the most superior overall performance.

The diffusion results for the test set of 2021 % Con EU-20 o EU-50?
reveal a significant dissimilarity in performance throughout the year. Figure~\ref{fig:image1} distinctly illustrates this dissimilarity for all three forecasted timeframes. Moving to Figure~\ref{fig:image2}, it becomes evident that this dissimilarity persists whether the prediction is generated using Single Diffusion, GDE (mean), or GDE (post-process). This consistency may be due to the intricacies inherent in precipitation forecasting, which can be especially demanding during specific periods of the year.
Precipitation patterns are notably prone to robust seasonal fluctuations~\cite{TUEL2022103855}. For example, shifts between seasons can trigger abrupt atmospheric changes, complicating the precise prediction of precipitation types and quantities~\cite{seasonal-variability}. Convective precipitation, which is more common during warm and humid conditions in the warmer months of the year, is closely linked with swiftly evolving weather systems like thunderstorms. It remains a formidable challenge for accurate forecasting, even in traditional operational meteorology~\cite{ray1986mesoscale, ConvectiveScaleWarnonForecastSystem}, as these systems manifest rapidly and are subject to an array of intricate and dynamic atmospheric processes.
% In letteratura sono stati proposti dei modelli specifici per la previsione di questo tipo di fenomeni (precipitazioni estreme - SWC) a livello di radar (aggiungere riferimenti?)
This led to the emergence in the literature of  proposals to tackle this specific task, namely severe weather  especially in regions more subject to these phenomena 
Additionally, the transit of weather fronts, encompassing cold and warm fronts, is more common during transitional seasons, such as spring and fall, when the differences in temperature and air masses are more pronounced~\cite{ahrens2006meteorology}. This phenomenon exerts swift alterations in precipitation distribution. The interplay between distinct air masses within these fronts introduces complexity, making the precise timing and location of precipitation a challenging problem.
% Il nostro modello, sfruttando la velocità del vento, dovrebbe essere in grado di affrontare questa complessità (stando al libro); ciò è forse visibile anche nelle figure, notando che nei mesi relativi all'autunno e soprattutto alla primavera, il modello non performa così male come in estate, dimostrando di soffrire fortemente dalle precipitazioni convettive. n.b. queste affermazioni mi sembrano un po' deboli, io forse non mi spingerei troppo

%\section{Conclusion and discussion} #strano mettere assieme la discussione con le conclusioni. Iolotre non c'è molta discussione
\section{Conclusions}
In this study, we tackled the challenging task of precipitation nowcasting using diffusion models. Through experimentation on well-established tasks documented in the literature, our proposed Generative Ensemble Diffusion (GED) approach achieved a greater overall performance with respect to competing U-Net based models. 

Our primary objective was to explore the hypothesis that a diffusion model could effectively capture the intricate chaotic behaviors of weather patterns by modeling its probability distribution. We worked on a subset of the ERA-5 dataset utilizing hourly data from a region of central Europe, training on the years 2016-2020 and testing on 2021. We trained our models on pre-processed versions of the dataset matching known works in literature~\cite{WF-UNet}.

Our experimental findings reveal that the incorporation of additional weather features enhances the prediction quality. Specifically, the inclusion of wind speed, land-sea mask, timestamp, and geopotential data contributed to improving the overall quality of our predictions.

Initially, our diffusion model's single generative predictions demonstrated a lower performance compared to a compatible prediction conducted using a U-net. However, a remarkable breakthrough emerged when we computed multiple generative predictions in parallel. By skillfully amalgamating these diverse outcomes through a post-processing step, we achieved a substantial improvement in the prediction quality, surpassing the performance of well-established U-net models. 

The probabilistic nature of the model, combined with its use of ensemble forecasting, makes it well-suited for forecasting rare events which occur with low probability but have a significant impact on the population and the economy~\cite{palmer2002economic}.

Overall, our work contributes to the advancement of precipitation nowcasting methodologies and offers a promising perspective on leveraging diffusion models to gain a deeper understanding of weather phenomena. The combination of GED with post-processing demonstrates the potential to enhance weather precipitation nowcasting, providing valuable information for various applications and decision-making processes.

A distinctive aspect of our work is that it has been conducted with very limited computing resources. The majority of computations were performed on a single workstation, which was equipped with a GPU Quadro RTXA4000 with 16GB of VRAM and 32GB of RAM.

Nevertheless, our research is part of an ongoing collaboration with the High Performance Computing Department of Cineca. As a next step, we intend to evaluate the model using the state-of-the-art Leonardo system, which offers significantly greater computational power. Specifically, our plan involves testing diffusion models on more intricate weather benchmarks. These benchmarks encompass predictions at varying spatial and temporal scales, with a particular focus on medium and long-term temporal ranges.

In addition, in the context of the European Cordis Project 
``Optimal High Resolution Earth System Models for Exploring Future Climate Changes'', we plan to apply our methodology to downscaling of meteorological indicators.
\bigskip

\subsection*{Acknowledgements}
This research was partially funded and supported by the following Projects:
\begin{itemize}
\item European Cordis Project ``Optimal High Resolution Earth System Models for Exploring Future Climate Changes'' (OptimESM),
Grant agreement ID: 101081193
\item Future AI Research (FAIR) project of the National Recovery and Resilience Plan (NRRP), Mission 4 Component 2 Investment 1.3 funded from the European Union - NextGenerationEU.
\item ISCRA Project ``AI for weather analysis and forecast'' (AIWAF)
\end{itemize}

\subsection*{Statements and Declarations}
The authors declare no competing interests.

\subsection*{Code availability}
The code relative to the presented work is archived in the following GitHub \href{repository}{https://github.com/fmerizzi/Precipitation-nowcasting-with-generative-diffusion-models}. The ERA-5 dataset used in our experiments can be openly accessed at the \href{Copernicus} {https://cds.climate.copernicus.eu} website.

\bibliography{bibliography,weather}

\end{document}